\documentclass[letterpaper, 10 pt, conference]{ieeeconf}  

\IEEEoverridecommandlockouts                              

\usepackage{amsmath}
\usepackage{amssymb}
\usepackage{mathtools}
\usepackage{graphicx}
\usepackage{caption}
\usepackage{booktabs}
\usepackage{multirow}
\usepackage{makecell}
\usepackage[normalem]{ulem}

\usepackage{thmtools}
\usepackage{amsfonts}
\usepackage{mathrsfs}
\usepackage{bm}
\usepackage{comment}
\usepackage{hyperref}

\usepackage[ruled,linesnumbered,vlined]{algorithm2e}
\SetAlFnt{\small}
\SetAlCapFnt{\small}
\SetAlCapNameFnt{\small}
\SetArgSty{textrm}
\SetInd{0.1em}{0.5em}
\SetCommentSty{emph}
\SetAlgoSkip{}
\usepackage{etoolbox}
\makeatletter
\patchcmd{\@algocf@start}
  {-1.5em}
  {0pt}
  {}{}
\makeatother

\title{\LARGE \bf
MFC-EQ: Mean-Field Control with Envelope $Q$-learning \\ for Moving Decentralized Agents in Formation%
\thanks{This work was supported by the NSERC under grant number RGPIN2020-06540 and a CFI JELF award.}}

\author{Qiushi Lin and Hang Ma%
\thanks{The authors are with the School of Computing Science, Simon Fraser University, Burnaby, BC, Canada {\tt\footnotesize \{qiushi\_lin, hangma\}@sfu.ca}}%
\thanks{Our code is publicly available at https://github.com/Qiushi-Lin/MFCEQ.}
}

\begin{document}

\maketitle
\thispagestyle{empty}
\pagestyle{empty}


\begin{abstract}

We study a decentralized version of Moving Agents in Formation (MAiF), a variant of Multi-Agent Path Finding aiming to plan collision-free paths for multiple agents with the dual objectives of reaching their goals quickly while maintaining a desired formation. The agents must balance these objectives under conditions of partial observation and limited communication. The formation maintenance depends on the joint state of all agents, whose dimensionality increases exponentially with the number of agents, rendering the learning process intractable. Additionally, learning a single policy that can accommodate different linear preferences for these two objectives presents a significant challenge. In this paper, we propose Mean-Field Control with Envelop $Q$-learning (MFC-EQ), a scalable and adaptable learning framework for this bi-objective multi-agent problem. We approximate the dynamics of all agents using mean-field theory while learning a universal preference-agnostic policy through envelop $Q$-learning. Our empirical evaluation of MFC-EQ across numerous instances shows that it outperforms state-of-the-art centralized MAiF baselines. Furthermore, MFC-EQ effectively handles more complex scenarios where the desired formation changes dynamically---a challenge that existing MAiF planners cannot address.
\end{abstract}


\section{Introduction}
Multi-Agent Path Finding (MAPF)~\cite{stern2019multi,ma2017ai} is a well-studied problem in various multi-agent systems, aiming to find collision-free paths for agents in a shared environment. Its applications include warehouse management~\cite{wurman2008coordinating}, airport surface operations~\cite{morris2016planning}, video games~\cite{ma2017feasibility}, and other multi-agent systems~\cite{gautam2012review}. Many of these applications require agents to adhere closely to a designated formation to accomplish collaborative tasks or ensure an efficient communication network. For example, in warehouse logistics, multiple robots or vehicles must collaborate to transport large objects, where maintaining a specific formation is critical for optimizing transport efficiency or ensuring reliable communication. Similarly, in video gaming or military strategy simulations, game characters or army personnel must move in formations to safeguard vulnerable members.

To tackle this challenge, \cite{li2020moving} formalized the bi-objective problem of Moving Agents in Formation (MAiF), which combines these two tasks, and proposed a centralized MAiF planner based on a leader-follower scheme and a search-based MAPF algorithm. However, existing MAiF planners are designed for centralized settings and are not applicable in practical scenarios where agents lack full observation of the environment. Additionally, centralized MAiF planners face significant computational challenges as the number of agents increases, making them unsuitable for real-time planning. Moreover, the only scalable MAiF planner, SWARM-MAPF~\cite{li2020moving}, lacks the flexibility to adjust to specific preferences between the two objectives, as it balances them only by setting a makespan allowance between two sets of heuristically determined waypoints, without guaranteeing optimization toward a targeted preference. We propose a novel approach for learning a general MAiF solver suitable for decentralized settings that can directly adapt to various preferences between the two objectives.

In the MAPF literature, reinforcement learning and imitation learning have been explored to solve MAPF in decentralized settings~\cite{sartoretti2019primal,liu2020mapper,ma2021distributed}. However, most learning-based MAPF solvers learn one homogeneous policy for any set of agents that treats nearby agents as part of the environment. This learning scheme does not seamlessly translate to decentralized MAiF. Unlike MAPF where the joint action cost can be directly decomposed into action costs of individual agents, formation maintenance in MAiF depends on the joint state of all agents at any given time. Each agent must not only avoid colliding with others but also coordinate with them to maintain proximity to the desired formation. The dimensionality of the joint state space grows exponentially with the number of agents, which hinders scalability. Additionally, trading off the two objectives under partial observation and limited communication further complicates this task.

In this paper, we formalize decentralized MAiF as a bi-objective multi-agent reinforcement learning problem. The major contributions of our work are as follows: We design a practical learning formalization for MAiF, including specifications for observations, actions, rewards, and inter-agent communication. To address the aforementioned challenges of MAiF, we propose a novel approach called \textsc{\textbf{M}ean \textbf{F}ield \textbf{C}ontrol with \textbf{E}nvelop $\bm{Q}$-learning} (MFC-EQ), a multi-agent reinforcement learning technique that optimizes toward any linear combination of two objectives for any number of agents, ensuring a stable and efficient learning process. MFC-EQ leverages mean-field control to approximate the collective dynamics of the agents, treating the interaction of each agent within the formation as influenced by the collective effect of others. This design choice facilitates seamless scalability to large-scale instances. Furthermore, MFC-EQ extends envelope $Q$-learning to a multi-agent setting, enabling the learning of a universal preference-agnostic model adaptable to any linear combination of the two objectives. To evaluate our method empirically, we extensively test MFC-EQ across various MAiF instances. Our results substantiate that MFC-EQ consistently produces solutions that surpass those generated by several centralized MAiF planners and scale effectively to large numbers of agents without long planning time. Additionally, the learned policy of MFC-EQ can directly adapt to more challenging tasks, including dynamically changing desired formations, which proves difficult for centralized MAiF planners.

\section{Problem Definition}
In this section, we first describe the standard MAiF formulation using terminology that facilitates the presentation of our learning approach. We then discuss how MAiF can be extended to a partially observable environment, which is a more practical problem setting. Finally, we define relevant concepts and outline the bi-objective optimization problem.

\subsection{Standard Formulation of MAiF}
In the standard formulation, an MAiF instance is defined on an undirected graph $G =(V, E)$ in a $d$-dimensional Cartesian system. Each location in $V$ can be identified by its coordinates $\bm{v} = (v_1, \dots, v_d) \in \mathbb{R}^d$. In this paper, superscripts represent agents' indices and boldface denotes multi-dimensional vectors. We define a set of $M$ agents $I = \{a^i | i \in [M]\}$, where $[M] = \{1, \dots, M\}$. Each agent has a unique start location $\bm s^i \in V$ and goal location $\bm g^i \in V$. Time is discretized, and at each time step, an agent can either wait at its current location or move from $\bm{v}$ to $\bm{v}'$, provided $(\bm{v}, \bm{v}') \in E$. We consider two types of collision between agents $a^i$ and $a^j$ at time step $t$: A vertex collision $\langle a^i, a^j, \bm{v}, t \rangle$ occurs when they occupy the same location $\bm{v}$, and an edge collision $\langle a^i, a^j, \bm{u}, \bm{v}, t \rangle$ occurs when $a^i$ moves from $\bm{u}$ to $\bm{v}$ while $a^j$ moves in the opposite direction.

The MAiF problem aims to find a set of $M$ collision-free paths $\Pi = \{\Pi^i | i \in [M]\}$ as a solution, where $\Pi^i = (\bm{p}_0^i, \dots, \bm{p}_{T^i}^i)$ represents agent $i$'s trajectory. Each solution is evaluated based on two objective functions, makespan and formation deviation. The makespan is defined as $T = \max_{1 \leq i \leq M} T^i$, representing the longest path length among all agents. The $\emph{formation}$ at time $t$ can be represented as an $M$-tuple, $\ell(t) = \langle \bm{p}^1(t), \dots, \bm{p}^M(t) \rangle$. The desired formation corresponds to a combination of all agents' goal locations, $\ell_{\bm{g}} = \langle \bm{g}^1, \dots, \bm{g}^M \rangle$. Following the definition in \cite{li2020moving}, the $\emph{formation deviation}$ between any two formations $\ell = \langle \bm{u}^1, \dots, \bm{u}^M \rangle$ and $\ell'= \langle \bm{v}^1, \dots, \bm{v}^M \rangle$ indicates the least effort required to transform $\ell$ into $\ell'$, defined as:
\begin{align}
 \mathscr{F} (\ell, \ell') 
&:= \min_{\bm{\Delta}} \sum_{i=1}^M \lVert \bm{u}^i - (\bm{v}^i + \bm{\Delta})\rVert_1 \nonumber \\
&= \sum_{i=1}^M \underbrace{\sum_{j=1}^d \lvert (\bm{u}_j^i - \bm{v}_j^i) - \bm{\Delta}_j \rvert}_{:= \mathscr{F}^i (\ell, \ell')},
\label{eq:formation_deviation}
\end{align}
where $j$ indexes the dimension for each vector and $\Delta_j = \emph{median}(\left\{\bm{u}_j^i - \bm{v}_j^i\right\}_{i \in [M]})$ is the median of the differences between the coordinates in the $j$-th dimension across all agents. The term $\mathscr{F}^i (\ell, \ell')$ denotes the component related to only agent $a^i$. We consider the average formation deviation per agent across all time steps, defined as $\mathscr{F}_{\emph{avg}} = \frac{1}{M} \sum_{t=0}^T \mathscr{F}(t)$, where $\mathscr{F}(t) = \mathscr{F}(\ell(t), \ell_{\bm{g}})$, which is more convenient in our decentralized setting than the total formation deviation used in \cite{li2020moving}. Additionally, we consider a linear combination of the two objectives:
\begin{equation*}
\mathtt{MIX}(\lambda) = \lambda T + (1 - \lambda) \mathscr{F}_{\emph{avg}},
\end{equation*}
where $\lambda$ balances the trade-off between makespan and formation deviation.
\begin{figure}[t]
\centering
\includegraphics[width=\columnwidth]{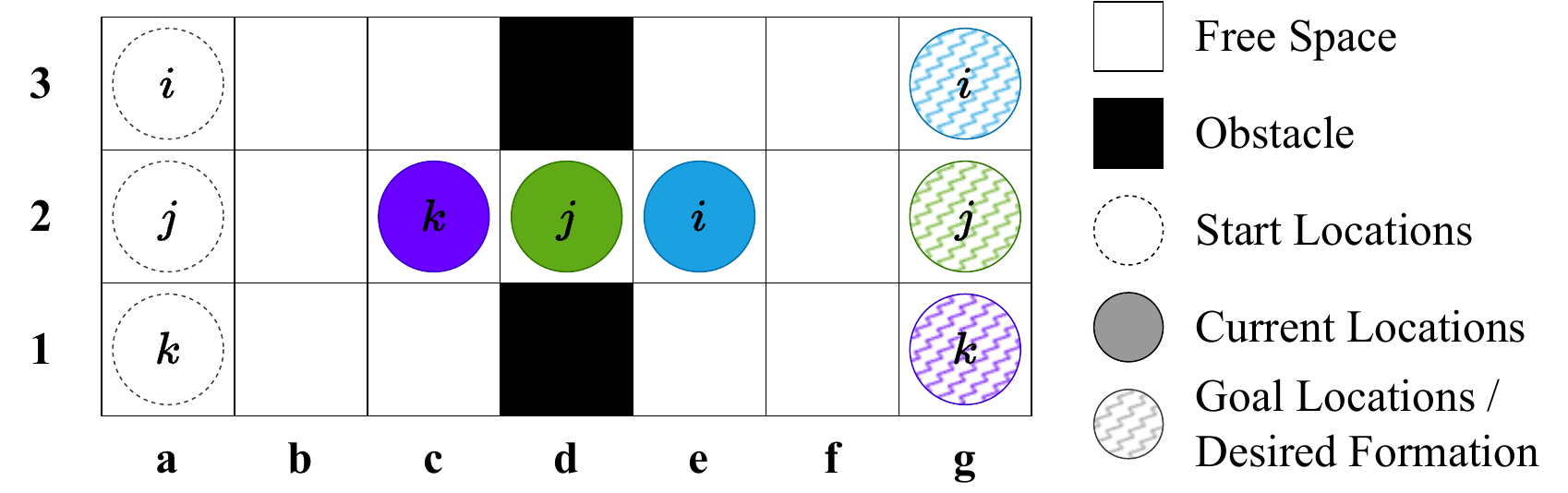}
\caption{Example of moving agents in formation.}
\label{fig:maif_example}
\end{figure}

\noindent\textbf{Example} A simple MAiF example is demonstrated in Fig.~\ref{fig:maif_example}. The start formation is $\langle a3, a2, a1 \rangle$ and the goal/desired formation is $\langle g3, g2, g1 \rangle$. The group of agents cannot go through the $d$ column while keeping the formation intact, so they have to change the formation. At the current time step $t$, the median position is $d2$ and the formation deviation is $\mathscr{F}(t) = \mathscr{F}^i(t) + \mathscr{F}^j(t) + \mathscr{F}^k(t) = 2 + 0 + 2 = 4$.

\subsection{Partially Observable Environment}
In this paper, we consider a more practical problem setting where, instead of having full knowledge of the environment, each agent can only observe part of its surroundings. We first introduce the standard decentralized partially observable Markov Decision Process (Dec-POMDP)~\cite{littman1994markov}. A Dec-POMDP is represented by a 7-tuple $\langle \mathcal{S}, \bm{A}, P_S, \bm{O}, P_O, R, \gamma \rangle$, where $\mathcal{S}$ is the global state space. $\bm{A} = \prod_{i=1}^{M} A^i $ and $\bm{O} = \prod_{i=1}^{M} S^i$, where $A^i$ and $S^i$ are agent $i$'s action and observation space. $P_S: \mathcal{S} \times \bm{A} \rightarrow \mathcal{S}$ describes the state-transition function, and $P_O: \mathcal{S} \times \bm{A} \rightarrow \bm{O}$ is the observation-transition function. $R$ is the reward function with the discount factor $\gamma$.
We adopt the standard Dec-POMDP framework to model our decentralized MAiF problem, assuming both the observation-transition and state-transition functions are deterministic. We also assume each agent can take an action based solely on its local observation and limited communication with others. Following existing learning-based MAPF literature~\cite{sartoretti2019primal,ma2021distributed}, we formalize this problem on a 2D 4-neighbor grid, though our method can be easily generalized to other settings. Partial observability restricts each agent's perception to a $\mathcal{L} \times \mathcal{L}$ square area centered on the agent, defined as its FOV. 

\subsection{Bi-Objective Optimization}
We now define the goal of this bi-objective optimization problem. Each MAiF solution is evaluated as $\bm{r} = (v, w)$, where $v$ denotes its makespan and $w$ denotes its average formation deviation per agent. We first define dominance: $\bm{r} = (v, w)$ dominates $\bm{r}' = (v', w')$, denoted as $\bm{r} \preceq \bm{r}'$, if and only if $v \leq v'$ and $w \leq w'$. A solution is Pareto-optimal if there does not exist any other solution that dominates it. The Pareto-optimal frontier is the set of all Pareto-optimal solutions. In the MAiF setting, we are also interested in evaluating each solution $r$ using a scalar function $f_{\bm{\omega}}(\bm{r}) = \bm{\omega}^{\top} \bm{r} $, where $\bm{\omega} \in \Omega$ represents a linear preference and $\Omega$ is the given distribution over a set of possible preferences. We use $\bm{\omega} = (\lambda, 1 - \lambda)^{\top}$, where $0 \leq \lambda < 1$.

\section{Related Work}
This section reviews related work on decentralized MAPF and MAiF, mean-field reinforcement learning, and multi-objective reinforcement learning.

\subsection{Learning-Based MAPF and MAiF Solvers}
Recently, reinforcement learning has been introduced to solve decentralized variants of MAPF~\cite{sartoretti2019primal,ma2021distributed,lin2023sacha}. These methods are designed to learn a decentralized model that can be generalized across different MAPF instances. Traditional centralized MAPF planners usually require full observation of the environment and must plan paths from scratch for each instance. In contrast, well-trained learning-based models offer the advantage of being applicable to any MAPF instances, regardless of the number of agents or the size of the environment.
For decentralized MAiF, \cite{liu2021moving} explored a similar setup to our work and proposed a hierarchical reinforcement learning scheme to divide the bi-objective task into unrelated sub-tasks. However, the hierarchical learning structure hinders the learned model's ability to adapt to different preference weights between the two objectives. Additionally, the simplistic network design struggles to scale to large numbers of agents. \cite{liu2021moving} also employs a different definition of formation deviation, and for that reason, we do not compare their results with our method in Section~\ref{sec:experiments}.

\subsection{Mean-Field Reinforcement Learning}
Inspired by mean-field theory~\cite{stanley1971phase} from physics, mean-field reinforcement learning has been proposed in~\cite{yang2018mean} to estimate the dynamics within an entire group of agents by modeling the interaction between each agent and the mean effect of all other agents as a whole. Since the dimensionality of the mean effect is independent of the number of agents, this method avoids the curse of dimensionality, providing a general framework for large-scale multi-agent tasks. This approach has been extended to partially observable stochastic settings~\cite{Srirampomfrl2021}, where certain distributions are used to sample agents' actions without the necessity of observing them. However, the sampling process is suited only for stochastic games, making it inapplicable to our task. 

\subsection{Multi-Objective Reinforcement Learning}
Multi-objective reinforcement learning methods can be categorized into three major types. Single-policy methods~\cite{gabor1998multi,mannor2001steering} convert a multi-objective problem into a single-objective optimization using linear or non-linear functions, but these methods cannot handle unknown preferences. Multi-policy methods~\cite{natarajan2005dynamic,parisi2014policy,van2014multi} work by updating a set of policies to approximate the true Pareto-optimal frontiers, but these methods require immense computational resources and are only feasible for problems with limited state and action spaces. Policy-adaptation methods either train a meta-policy that adapts to different preferences on the fly~\cite{chen2019meta} or learn a policy conditioned on different preference weights~\cite{castelletti2011multi,abels2019dynamic,yang2019generalized}. Envelop $Q$-learning~\cite{yang2019generalized} has been proposed to increase sample efficiency by introducing a novel envelop operator for updating the multi-objective $Q$-function, which has become a standard approach for tackling multi-objective problems with linear preferences.

\section{MFC-EQ}
This section presents the design of our learning framework for decentralized MAiF. We first outline the learning environment, including agents' observation, communication, action, and reward functions. We then elaborate on the bi-objective multi-agent learning process based on mean-field theory and envelope $Q$-learning.

\subsection{Environment and Model Design}

\subsubsection{Observation}
As with most research in the MAPF community~\cite{stern2019multi}, we study our problem in a 2D 4-neighbor grid environment. To mimic many real-world robotics applications where robots have limited visibility and sensing range, each agent in our grid world can observe only its field of view (FOV), represented by its surrounding $\mathcal{L} \times \mathcal{L}$ area. Each agent's observation is represented by multi-channel feature maps. The first few channels indicate obstacles and other neighboring agents' positions. Inspired by some decentralized MAPF solvers~\cite{liu2020mapper,ma2021distributed}, the rest of the channels encompass heuristic information, where each cell in the FOV is assigned a value proportional to the short-path distance from that cell to the agent's goal.

\subsubsection{Action}
Agents can move to their cardinally adjacent cells at each time step. The action taken by agent $i$ at time step $t$, denoted by $a^i_t \in \mathbb{R}^5$, is a 5-dimensional one-hot vector where each dimension represents one of the actions: $\{up, down, left. right, wait\}$. The first four actions move the agent to a new cell, shifting its observation accordingly. The $wait$ action keeps the agent in its current cell, which is especially crucial for formation control, allowing other agents to catch up and reduce formation deviation.

\subsubsection{Multi-Agent Communication}
To maintain the desired formation, agents need to not only communicate with nearby agents within their FOVs but also with those outside. We specifically design communication messages to convey critical information under low communication bandwidth.

In many real-world robotics applications, each agent can only access pairwise relative positions between itself and other agents. Assume that the current formation at time step $t$ is $\ell_{\bm{p}} = \langle \bm{p}^1, \dots, \bm{p}^M \rangle$ and the desired formation is $\ell_{\bm{g}} = \langle \bm{g}^1, \dots, \bm{g}^M \rangle$. The relative position between agent $i$ and $j$ is defined as $\bm{p}^{i,j} = \bm{p}^j - \bm{p}^i$ (resp., $\bm{g}^{i,j}$). Agent $i$ receives $\{\bm{p}^{i,j}\}_{j \in [M]}$ in real-time and knows the relative positions in the goal formation, $\{\bm{g}^{i,j}\}_{j \in [M]}$, pre-calculated before execution. With this information, even without knowing its absolute position, an agent can still calculate the formation deviation. As defined in Eq.~\eqref{eq:formation_deviation}, 
$\mathscr{F} (\ell_{\bm{p}}, \ell_{\bm{g}}) = \min_{\bm{\Delta}} \sum_{m=1}^M \lVert \bm{p}^m - (\bm{g}^m + \bm{\Delta})\rVert_1 = \sum_{m=1}^M \sum_{n=1}^d \lvert (\bm{p}_n^m - \bm{g}_n^m) - \bm{\Delta}_n \rvert$ where $\bm{\Delta}_n$ is the median of $\{\bm{p}_n^m - \bm{g}_n^m\}_{m \in [M]}$. Recall that $d$ is the dimension of agents' coordinates. It is easy to verify that
\begin{equation*}
\mathscr{F} (\ell_{\bm{p}}, \ell_{\bm{g}}) = \sum_{m=1}^M \sum_{n=1}^d \lvert (\bm{p}_n^m - \bm{g}_n^m - \bm{C}_n) - \bm{\Delta}'_n \rvert,
\end{equation*}
where $\bm{C}$ is any constant $d$-dimensional vector and $\bm{\Delta}'_n$ is the median of $\{\bm{p}_n^m - \bm{g}_n^m - \bm{C}_n\}_{m \in [M]}$, as all the values and the median are shifted by the same margin. Substituting $\bm{C}$ with $\bm{p}^i - \bm{g}^i$, we can rewrite the formation deviation only using relative positions:
\begin{align*}
&\sum_{m=1}^M \sum_{n=1}^d \lvert (\bm{p}_n^m - \bm{g}_n^m) - \bm{\Delta}_n \rvert \\
=& \sum_{m=1}^M \sum_{n=1}^d \lvert [(\bm{p}_n^m - \bm{p}_n^i) - (\bm{g}_n^m - \bm{g}_n^i)] - \bm{\Delta}^{*}_n \rvert \\
=& \sum_{m=1}^M \sum_{n=1}^d \lvert (\bm{p}_n^{i,m} - \bm{g}_n^{i,m} - \bm{\Delta}^{*}_n \rvert,
\end{align*}
where $\bm{\Delta}^{*}_n$ is the median of $\{\bm{p}_n^{i,m} - \bm{g}_n^{i,m}\}_{m \in [M]}$. Therefore, agent $i$ can calculate the formation deviation based on only the relative positions with time complexity $\mathcal{O}(d \cdot M)$. Additionally, agents can easily infer the mean action based on relative positions.

It is worth mentioning that allowing agents to communicate relative positions does not make the problem centralized. Each agent remains unaware of the surrounding environment of agents outside its FOV and cannot predict their observations or next actions based solely on relative positions.

\subsubsection{Reward}
The reward function for agent $i$ after taking action $a$ at time step $t$, $\bm{r}_t^i(s_t^i, a_t^i) \in \mathbb{R}^2$, is a 2-tuple. The first element concerns the makespan. We adopt the individual cost function from DHC~\cite{ma2021distributed}: The moving cost of agent $i$ at time step $t$ with action $a^i$ is
\begin{equation*}
 c_t^i(s^i, a^i) =
\begin{cases}
-0.075 & \text{collision-free $a^i$} \\
-0.5 & \text{collision (with obstacles or agents)} \\
3 & \text{reach goal} \\
0 & \text{stay on goal}
\end{cases}.
\end{equation*}
Collision-free actions, including moves ($up$, $down$, $left$, or $right$) and $wait$ (on or away from the goal), are slightly penalized to encourage agents to reach their goals quickly. The second element concerns the formation deviation. As defined in Eq.~\eqref{eq:formation_deviation}, we add the individual portion of the collective formation deviation that is dedicated to agent $j$, namely $\mathscr{F}_t^j (\ell_t, \ell^g)$. We negate the formation deviation to minimize it by maximizing rewards. Thus, the reward function is:
\begin{equation}
\bm{r}_t^j(s_t^j, a_t^j) = (c_t^j, -\mathscr{F}_t^j (\ell_t, \ell^g))^\top.
\label{eq:reward}
\end{equation}

\begin{figure}[t]
\centering
\includegraphics[width=\columnwidth]{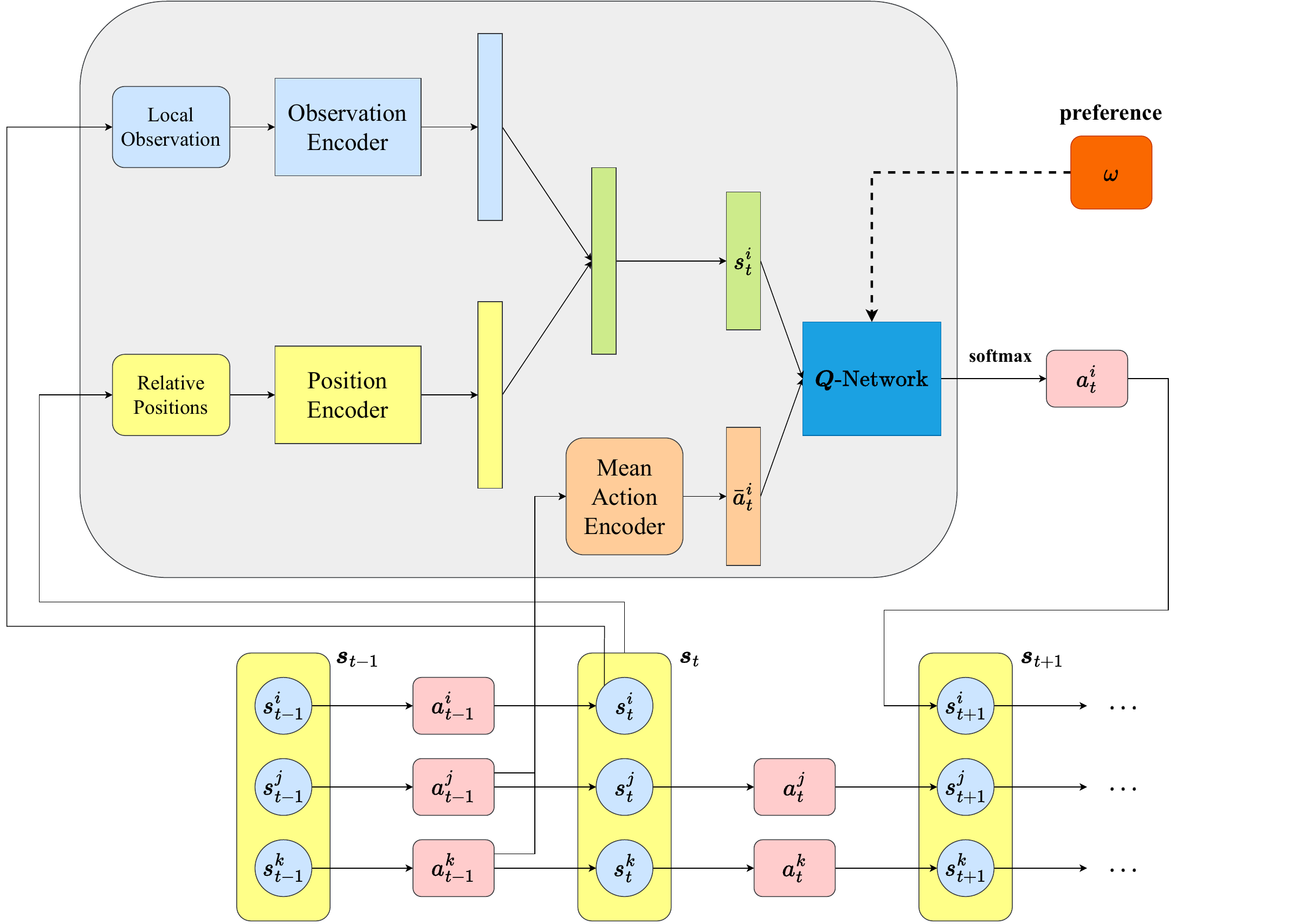}
\caption{Illustration of the model architecture of MFC-EQ. The bottom demonstrates the state/observation transition in the partially observable environment. The agent's $Q$-network gathers information from the environment through partial observation and limited communication and chooses the next action accordingly.}
\label{fig:model_architecture}
\end{figure}

\subsubsection{Model Architecture}
Given the partially observable multi-agent environment, we design a $Q$-network, with the learning algorithm introduced later. As shown in Fig.~\ref{fig:model_architecture}, the network projects each agent's observations and communication messages into a corresponding action. The local observation and relative positions are fed into two separate encoders: The encoders use stacked convolution layers followed by multiple linear layers, and these encodings are then concatenated and passed through another set of linear layers to obtain the final state representation, $s^i_t$, for agent $i$ at time step $t$. The mean action is calculated by collecting other agents' actions from the previous time step. Lastly, the stacked linear layers project these inputs to $Q$-values, conditioning on the state, action, mean action, and given preference. The agent selects its next action that maximizes the $Q$-function produced by the $Q$-network.

\subsection{Mean-Field and Envelop Optimality}
\label{sec:mfeo}
In the following, we discuss the learning algorithm in detail. Learning multiple policy networks, $\pi = [\pi^1, \dots, \pi^M]$, for this bi-objective multi-agent task is highly challenging. To simplify this, we make some common assumptions.

\subsubsection{Mean-Field Approximation}
The goal of MAiF is to minimize the makespan and formation deviation. With our specifically designed reward function, the return---the discounted sum of all rewards from the initial to the goal joint state, $\sum_t\sum_j\gamma^t\bm{r}_t^j(\bm{s}_t, \bm{a}_t)$---reflects the true values of the two objectives. Therefore, the learning goal is to find a set of policies that maximize the general sum of $Q$-values $\arg \max_{\pi^1, \dots, \pi^M} \sum_{j=1}^M \bm{\omega}^{\top} \bm{Q}^{\pi^j}(s^j, \bm{a})$ given the linear preference $\bm{\omega}$. However, the dimensionality of $\bm{s}$ and $\bm{a}$ grows exponentially with the number of agents, rendering efficient learning infeasible.
To address this, we introduce mean-field reinforcement learning. Similar to \cite{yang2018mean}, we first adopt the common assumptions of homogeneity and locality. Homogeneity assumes that each agent shares the same policy, so $\pi^i = \pi^j$ for all $i \neq j$. Locality, derived from partial observability, suggests that agents' actions depend only on their visible surroundings.
Assuming actions are represented by one-hot vectors, we define the mean action as:
\begin{equation}
\bar{a}^j_t = \frac{1}{|\mathcal{N}^j|} \sum_{k \in \mathcal{N}^j} a^k_t,\; a^k_t \sim \pi^k(\cdot | s^k, \bar{a}^k_{t-1}),
\label{eq:mean_action}
\end{equation}
where $\mathcal{N}^j$ denotes agent $j$'s neighboring agents and $\pi^j$ represents its policy. With homogeneity and locality, under a certain preference $\bm{\omega}$, the local pairwise interactions can be approximated by the interplay of each agent with the mean effect from its neighbors:
\begin{align}
\bm{\omega}^{\top} \bm{Q}(s_t^j, \bm{a}_t) &= \frac{1}{|\mathcal{N}^j|} \sum_{k \neq j} \bm{\omega}^{\top} \bm{Q}(s_t^j, a_t^j, a_t^k) \nonumber \\
&= \bm{\omega}^{\top} \bm{Q}(s_t^j, a_t^j, \bar{a}_t^j), \label{eq:mean_field}
\end{align}
where $\bm{a}_t$ is the joint action, $a_t$ is the single-agent action, and $\bar{a}_t$ is the mean action. Given this approximated $Q$-function, we derive the agent's policy function using the softmax parameterization with the Boltzmann parameter $\beta$:
\begin{equation}
 \pi^j(a_t^j | s_t^j, \bar{a}^j_{t-1}) = \frac{\exp(\beta\, \bm{\omega}^{\top} \bm{Q}(s_t^j, a_t^j, \bar{a}^j_{t-1}))}{ \sum_{a \in A^j}\exp(\beta\,  \bm{\omega}^{\top} \bm{Q}(s_t^j, a, \bar{a}^j_{t-1}))}.
\label{eq:policy}
\end{equation}

\subsubsection{Bellman Optimality Operator}
To extend this framework to multi-objective reinforcement learning, we modify the envelop $Q$-learning~\cite{yang2019generalized} by combining the mean-field operator with the envelop optimality operator. We first condition all $Q$-values on the linear preference $\bm{ \omega}$, as in $\bm{Q}(\bm{s}, \bm{a}, \bm{\omega})$. Similar to standard $Q$-learning~\cite{watkins1992q}, the bi-objective multi-agent Bellman optimality operator $\mathcal T$ is defined as:
\begin{align}
&(\mathcal{T} \bm{Q})(\bm s_t, \bm a_t, \bm \omega) := \sum_{j=1}^{M} \bm r^j(s_t^j, a_t^j) +  \nonumber \\
& \gamma \, \mathbb{E} \left[\arg_Q \left\{ \max_{\bm{\omega}' \in \Omega} \sum_{j=1}^{M} \max_{a^j \in A^j}\bm{\omega}^{\top} \bm{Q} (s_{t+1}^j, a^j, \bar{a}_{t+1}^j,\bm{\omega}') \right\} \right],
\label{eq:operator}
\end{align}
where $\arg_Q$ takes the maximized bi-objective $Q$-values under the preference $\bm{\omega}$. This operator mirrors the Bellman optimality operator in standard $Q$-learning for single-agent RL and provides the temporal difference (TD) target. By maximizing $\bm{\omega}'$ over the next state and its onward trajectory, this approach offers an optimistic perspective on its future rewards. Iteratively applying this operator to the $Q$-function allows for convergence to a near-optimal $Q$-function~\cite{yang2019generalized}.

\subsection{Double $Q$-learning}
Off-policy RL algorithms allow for greater exploration in the state-action space and have been adopted by several learning-based MAPF solvers (e.g., \cite{ma2021distributed}). We thus design our learning algorithm based on the double $Q$-learning~\cite{hasselt2010double} with two different loss functions. Algorithm~\ref{algo:mfc-eq} presents the detailed learning framework. In the rollout phase (Line \ref{line:rollout_start}-\ref{line:rollout_end}), we sample the transitions in the multi-agent environment with homogeneous policies. Once enough transitions are collected in the replay buffer, the learning phase begins (Line \ref{line:learning_start}-\ref{line:learning_end}). Given a mini-batch of $N$ transitions and $N_{\bm{\omega}}$ preferences, the TD target $\bm{y} = (\mathcal{T} \bm{Q})(\bm s, \bm a, \bm \omega)$ is estimated via Eq.~\eqref{eq:operator}. The first loss function is computed as the $L_2$-norm of the multi-objective TD:
\begin{equation*}
 L_A(\theta) = \mathbb E_{\bm{s}, \bm{a}, \bm{\omega}} \biggl[ \| \bm{y} - \sum_{j=1}^M\bm{Q}_{\theta}(s^j, a^j, \bar{a}^j, \bm{\omega}) \|_2^2 \biggr].
\end{equation*}
Although this loss function closely estimates the true expected return, its non-smooth surface makes the early learning steps challenging. We combine this with an additional loss function using the projected TD:
\begin{equation*}
 L_B(\theta) = \mathbb E_{\bm{s}, \bm{a}, \bm{\omega}} \biggl[ \lvert \bm{\omega}^{\top} (\bm{y} - \sum_{j=1}^M\bm{Q}_{\theta}(s^j, a^j, \bar{a}^j, \bm{\omega})) \rvert \biggr].
\end{equation*}
$L_A(\theta)$ provides a closer estimation of the true $Q$-function, while $L_B(\theta)$ smooths the optimization landscape. We train the $Q$-network using homotopy optimization~\cite{watson1989modern} based on the combination of these two loss functions:
\begin{equation}
L(\theta) = (1 - \zeta) L_A(\theta) + \zeta L_B(\theta),
\label{eq:loss}
\end{equation}
where, in our case, we gradually increases $\zeta$ from $0$ to $1$ as learning progresses.

\begin{algorithm}[t]
\linespread{0.42}\selectfont
\caption{Mean-Field Control with Envelop $Q$-learning}
\label{algo:mfc-eq}
Initialize the $Q$-network $\bm{Q}_{\theta}$ and the target $Q$-network $\bm{Q}_{\bar{\theta}}$ \\
Initialize the replay buffer $\mathcal D$ and set $\zeta = 0$ \\
\For{episode = $1, \dots, E$} {
    Initialize $\bar{a}_0^j$ for all $j \in [M]$ \\
    \For{$t=1, \dots, T_{\emph{max}}$} {
        Sample $\bm{\omega} \sim \Omega$ and then $a_t^j$ from Eq.~\eqref{eq:policy} \label{line:rollout_start} \\
        Compute new mean actions $\bar{a}_t^j$ by Eq.~\eqref{eq:mean_action} for all $j \in [M]$ \\
        Take the joint action $\textbf{a}_t = [a_t^1, \dots, a_t^M]$ from the state $\textbf{s}$ to the next state $\textbf{s}_{t+1}$ \\
        Compute the reward $\textbf{r}_t = [r^1,\dots,r^M]$ by Eq.~\eqref{eq:reward} \\
        Store the transition, $\langle \bm{s}_t,\bm{a}_t,\bm{r}_t,\bm{s}_{t+1},\bar{\bm{a}} \rangle$, into $\mathcal{D}$, where $\bar{\bm{a}}_t = [\bar{a}_t^1,\dots,\bar{a}_t^M]$ is the collection of mean actions \\
    } \label{line:rollout_end}
    \If{update} {\label{line:learning_start}
        Sample $N$ transitions from $\mathcal{D}$ and $N_{\bm{\omega}}$ preferences from $\Omega$ \\
        Compute the TD target using the operator in Eq.~\eqref{eq:operator} \\
        Update $Q_\theta$ by minimizing the loss from Eq.~\eqref{eq:loss} \\
    } \label{line:learning_end}
    Update $\bm{Q}_{\bar{\theta}}$ with the learning rate $\alpha$: $\bar{\theta} \gets \alpha \theta + (1 - \alpha)\bar{\theta}$ \\
    Increase $\zeta$ along the predefined homotopy path \\
}
\end{algorithm}

\section{Empirical Evaluation} \label{sec:experiments}
This section presents our experimental results conducted on a 2.3GHz Intel Xeon server with 8 NVIDIA A40 GPUs.

\subsection{Experimental Setups}
We use 4-neighbor grids without placing obstacles in the top-left and bottom-right corners. The default obstacle density in the remaining areas is set to $10\%$. Agents start at the top-left corner and move toward the bottom-right corner. Following existing learning-based MAPF literature, the FOV size is set to $9 \times 9$. The formation in the goal position defines the desired formation. For each data point in the results, we average the outcomes over $100$ samples, generated from about $10$ random maps and about $10$ random formations.

\begin{figure}[t]
\centering
\includegraphics[width=\columnwidth]{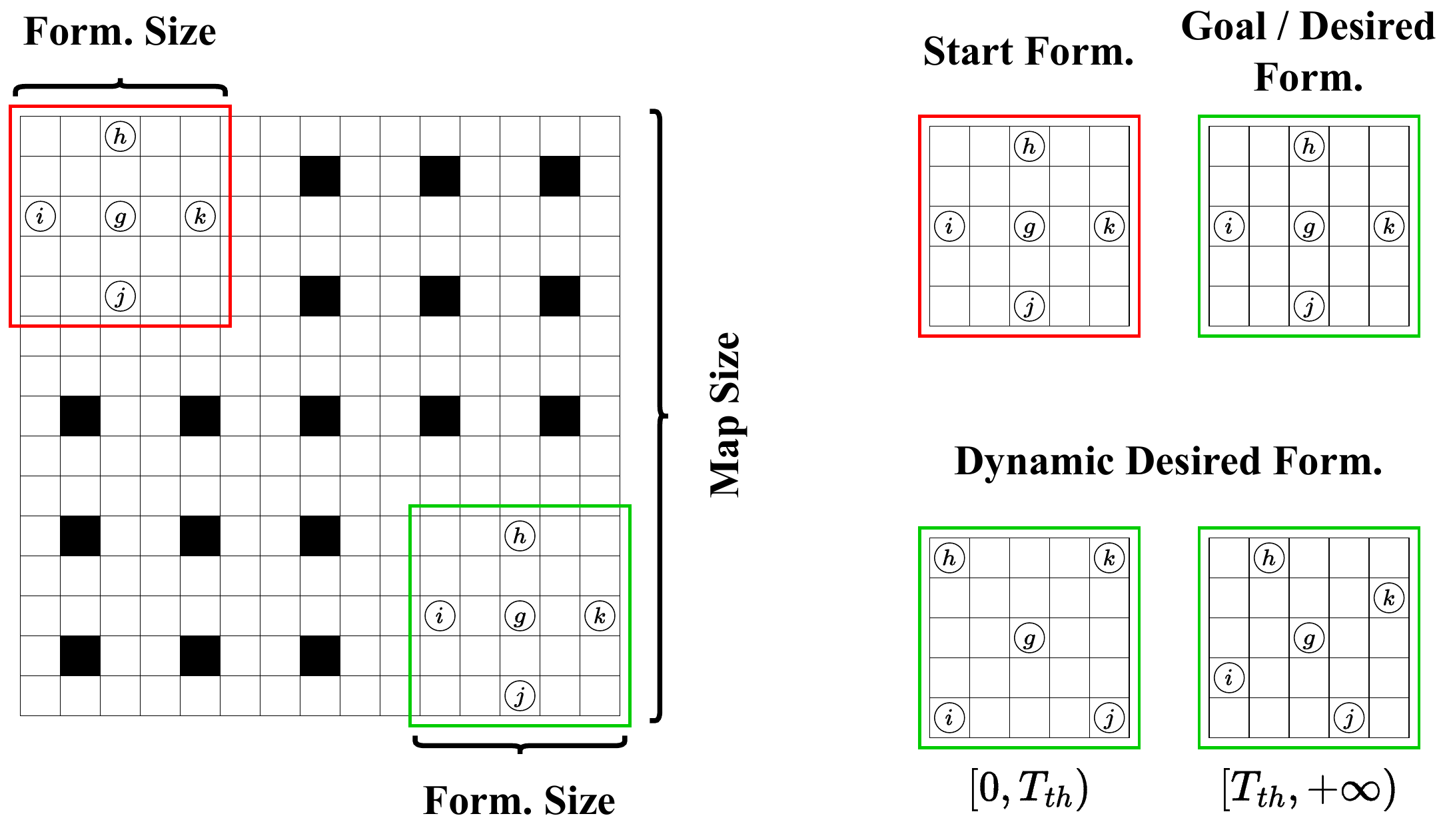}
\caption{Demonstration of experiment environments.}
\label{fig:experimental_setups}
\end{figure}

\subsection{Training of MFC-EQ}
Similar to \cite{ma2021distributed}, we employ the scheme of curriculum learning~\cite{bengio2009curriculum} to gradually introduce more challenging tasks into the training data. For each formation size, the training starts with a $30 \times 30$ map and $5$ agents, and we will increase either the map size or the number of agents once the success rate exceeds $0.9$. The final model was trained for $500,000$ episodes with a batch size of $192$. 

\subsection{Centralized Baselines}
We compare our method against the following centralized planning methods.

\subsubsection{Scalarized Prioritized Planning (SPP)}
Since solving MAiF optimally is NP-hard, we developed an efficient yet suboptimal baseline based on the prioritized planning algorithm~\cite{silver2005cooperative}. Each agent is given a unique priority, and paths are planned sequentially using a low-level A* search that respects the paths of higher-priority agents. The A* search uses a scalarized $f$-value combining the makespan $f$-value and the formation deviation $f$-value. For any node $n$, we define its cost from the start node to node $n$ as $T_n$, and the scalarized $f$-value can be written as:
\begin{equation*}
f(n) = \lambda  \vphantom{\sum_{x}} \bigl[T_n + \emph{dist}(\bm{v}^i_{T_n}, \bm{g}^i)\bigr] + (1 - \lambda) \sum_{t=1}^{T_n}\mathscr{F}^i_t(\bm{v}^i_t, \bm{g}^i),
\end{equation*}
where $\emph{dist}(\cdot, \cdot)$ is the Manhattan distance and $\mathscr{F}^i_t$ represents the partial formation deviation involving agent $i$ and higher-priority agents at time step $t$ as defined in Eq.~\ref{eq:formation_deviation}. Although this baseline is incomplete, it can often find a solution quickly. However, the quality, especially for formation deviation, may suffer in congested environments with many agents. Unlike SWARM-MAPF, this planner can target any given linear preference. We tested its performance by varying $\lambda$ from $\{0.1, 0.3, 0.5, 0.7, 0.9\}$.
 
\subsubsection{SWARM-MAPF (SWARM)}
SWARM-MAPF~\cite{li2020moving} is a state-of-the-art centralized method that integrates swarm-based formation control with a MAPF algorithm. It operates in two phases: In Phase 1, SWARM-MAPF first calculates the lower bound of the makespan $B = \max_{1 \leq i \leq M} \emph{dist}(s_i, g_i)$ and then selects a leader to plan a path of length bounded by the user-specified parameter $w$ multiplied by $B$, ensuring that the path is sufficiently distant from obstacles to allow other agents to preserve formation while avoiding obstacles. In Phase 2, SWARM-MAPF employs a modified conflict-based search~\cite{sharon2015conflict} to replan critical segments while minimizing the makespan. While SWARM-MAPF is complete, it cannot specifically target a given preference, as the trade-off between objectives cannot be directly controlled through the parameter $w$. 

\subsubsection{Joint State A* (JSA*)}
Joint state A*~\cite{pearl1982studies} applies the $\epsilon$-constraint search algorithm~\cite{haimes1971bicriterion} directly in the joint state space. The joint state assigns all agents a set of different locations. The operator assigns each agent a set of non-colliding move or wait actions. The OPEN list sorts nodes by makespan, while the FOCAL list breaks ties based on formation deviation~\cite{li2020moving}. Due to the exponential growth of the joint state space with the number of agents, this method is only feasible for small instances (fewer than $5$ agents in our setups). Varying $\epsilon$ in the focal search guarantees finding the Pareto-optimal frontier.

\subsection{Main Results}

\begin{table}[t]
\centering
\caption{Results of MFC-EQ with different preferences evaluated by different scalarized objectives.}
\Huge\scalebox{0.26}{%
\begin{tabular}{c|rr|r|r|r|r|r}
\hline
$\bm{\omega}(\lambda)$ & \multicolumn{1}{c}{\begin{tabular}[c]{@{}c@{}}Make-\\ span\end{tabular}} & \multicolumn{1}{c|}{\begin{tabular}[c]{@{}c@{}}Form.\\ Dev.\end{tabular}} & \multicolumn{1}{c|}{$\mathtt{MIX}(0.1)$} & \multicolumn{1}{c|}{$\mathtt{MIX}(0.3)$} & \multicolumn{1}{c|}{$\mathtt{MIX}(0.5)$} & \multicolumn{1}{c|}{$\mathtt{MIX}(0.7)$} & \multicolumn{1}{c}{$\mathtt{MIX}(0.9)$} \\ \hline
0.1 & 106.33 & 14.67 & \textbf{23.84} & 42.17 & 60.50 & 78.83 & 97.16 \\
0.3 & 101.14 & 15.37 & 23.95 & \textbf{41.10} & 58.26 & 75.41 & 92.56 \\
0.5 & 98.64 & 16.84 & 25.02 & 41.38 & \textbf{57.74} & 74.10 & 90.46 \\
0.7 & 96.74 & 19.16 & 26.92 & 42.43 & 57.95 & \textbf{73.47} & 88.98 \\
0.9 & 96.42 & 21.75 & 29.22 & 44.15 & 59.09 & 74.02 & \textbf{88.95} \\ \hline
\end{tabular}%
}
\label{tab:preference}
\end{table}

\subsubsection{Linear Preferences}
We first evaluate the ability of our learned $Q$-network to adapt to different preferences using an environment with $16$ agents and a $9 \times 9$ formation size on a $48 \times 48$ map. We test different preferences $\bm{\omega} = (\lambda, 1 - \lambda)^{\top}$ by varying $\lambda$ from $0.1$ to $0.9$. We also evaluate each result under different $\mathtt{MIX}(\lambda)$ objectives by varying $\lambda$ from the same set of values. Table~\ref{tab:preference} shows $5$ different solutions, with each $\mathtt{MIX}$ column highlighting the solution that minimizes the projection onto that particular preference. We observe that each $\mathtt{MIX}(\lambda)$ objective is minimized when the corresponding preference $\bm{\omega}(\lambda)$ is fed into the $Q$-network. The results suggest that the $Q$-network trained with MFC-EQ can adapt to different preferences, producing multiple solutions tailored to each given preference.

\subsubsection{Numbers of Agents}
We evaluate MFC-EQ with different numbers of agents across different map sizes and compare the results with centralized baselines. For SPP and MFC-EQ, $\lambda$ is set to $0.5$. For SWARM, $w$ is set to $1.0$. The runtime limit is $30$ seconds for MFC-EQ and $5$ minutes for SWARM and SPP. As shown in Table~\ref{tab:num_agents}, MFC-EQ does not always achieve perfect success rates due to the partially observable environment but maintains relatively high and acceptable rates. It generally achieves slightly lower success rates than SWARM for large numbers of agents, despite having a much shorter runtime limit. MFC-EQ often outperforms SWARM when evaluated under the given preference. These results suggest that MFC-EQ scales well to large numbers of agents across different map sizes.

\subsubsection{Formation Sizes}
We repeat the above experiment with different formation sizes, using various obstacle-free corner sizes and randomly generated desired formations. The larger the corner, the more spread out the formation tends to be. The number of agents is fixed at $16$. As shown in Table~\ref{tab:form_size}, smaller formations are generally more challenging, resulting in larger makespans and formation deviations. Compared to SWARM, SPP typically achieves a better makespan but much worse formation deviation. MFC-EQ outperforms the two baselines with respect to both objectives in most cases.

\begin{table}[t]
\centering
\caption{Results for MFC-EQ and centralized baselines with different numbers of agents in various sizes of grids.}
\Huge\scalebox{0.26}{%
\begin{tabular}{c|c|rrr|rrr|rrr|rrr}
\hline
 &  & \multicolumn{3}{c|}{Success Rate} & \multicolumn{3}{c|}{Makespan} & \multicolumn{3}{c|}{Form. Dev.} & \multicolumn{3}{c}{$\mathtt{MIX}(0.5)$} \\ \hline
\begin{tabular}[c]{@{}c@{}}Map\\ Size\end{tabular} & $M$ & \multicolumn{1}{c|}{\rotatebox[origin=c]{90}{\parbox[c]{2.3cm}{\centering SPP}}} & \multicolumn{1}{c|}{\rotatebox[origin=c]{90}{\parbox[c]{2.3cm}{\centering SWA- RM}}} & \multicolumn{1}{c|}{\rotatebox[origin=c]{90}{\parbox[c]{2.3cm}{\centering MFC- EQ}}} & \multicolumn{1}{c|}{\rotatebox[origin=c]{90}{\parbox[c]{2.3cm}{\centering SPP}}} & \multicolumn{1}{c|}{\rotatebox[origin=c]{90}{\parbox[c]{2.3cm}{\centering SWA- RM}}} & \multicolumn{1}{c|}{\rotatebox[origin=c]{90}{\parbox[c]{2.3cm}{\centering MFC- EQ}}} & \multicolumn{1}{c|}{\rotatebox[origin=c]{90}{\parbox[c]{2.3cm}{\centering SPP}}} & \multicolumn{1}{c|}{\rotatebox[origin=c]{90}{\parbox[c]{2.3cm}{\centering SWA- RM}}} & \multicolumn{1}{c|}{\rotatebox[origin=c]{90}{\parbox[c]{2.3cm}{\centering MFC- EQ}}} & \multicolumn{1}{c|}{\rotatebox[origin=c]{90}{\parbox[c]{2.3cm}{\centering SPP}}} & \multicolumn{1}{c|}{\rotatebox[origin=c]{90}{\parbox[c]{2.3cm}{\centering SWA- RM}}} & \multicolumn{1}{c}{\rotatebox[origin=c]{90}{\parbox[c]{2.3cm}{\centering MFC- EQ}}} \\ \hline
\multirow{3}{*}{\parbox[c]{0.7cm}{\centering $32$ $\times$ $32$}} & 10 & \multicolumn{1}{r|}{1.00} & \multicolumn{1}{r|}{1.00} & 1.00 & \multicolumn{1}{r|}{48.30} & \multicolumn{1}{r|}{59.32} & 60.24 & \multicolumn{1}{r|}{29.79} & \multicolumn{1}{r|}{6.07} & 4.35 & \multicolumn{1}{r|}{39.05} & \multicolumn{1}{r|}{32.70} & \textbf{32.30} \\
 & 20 & \multicolumn{1}{r|}{1.00} & \multicolumn{1}{r|}{0.99} & 0.99 & \multicolumn{1}{r|}{49.03} & \multicolumn{1}{r|}{63.17} & 60.38 & \multicolumn{1}{r|}{32.42} & \multicolumn{1}{r|}{12.38} & 10.04 & \multicolumn{1}{r|}{40.73} & \multicolumn{1}{r|}{37.78} & \textbf{35.21} \\
 & 30 & \multicolumn{1}{r|}{0.79} & \multicolumn{1}{r|}{0.96} & 0.90 & \multicolumn{1}{r|}{51.54} & \multicolumn{1}{r|}{59.09} & 54.59 & \multicolumn{1}{r|}{42.25} & \multicolumn{1}{r|}{20.64} & 20.32 & \multicolumn{1}{r|}{46.90} & \multicolumn{1}{r|}{39.87} & \textbf{37.46} \\ \hline
\multirow{3}{*}{\parbox[c]{0.7cm}{\centering $48$ $\times$ $48$}} & 10 & \multicolumn{1}{r|}{1.00} & \multicolumn{1}{r|}{0.99} & 0.99 & \multicolumn{1}{r|}{80.44} & \multicolumn{1}{r|}{98.10} & 88.07 & \multicolumn{1}{r|}{53.91} & \multicolumn{1}{r|}{8.18} & 11.05 & \multicolumn{1}{r|}{67.18} & \multicolumn{1}{r|}{53.14} & \textbf{49.56} \\
 & 20 & \multicolumn{1}{r|}{0.95} & \multicolumn{1}{r|}{0.99} & 0.96 & \multicolumn{1}{r|}{82.07} & \multicolumn{1}{r|}{108.84} & 104.28 & \multicolumn{1}{r|}{70.44} & \multicolumn{1}{r|}{23.70} & 21.49 & \multicolumn{1}{r|}{76.26} & \multicolumn{1}{r|}{66.27} & \textbf{62.89} \\
 & 30 & \multicolumn{1}{r|}{0.74} & \multicolumn{1}{r|}{0.94} & 0.88 & \multicolumn{1}{r|}{84.92} & \multicolumn{1}{r|}{101.52} & 107.42 & \multicolumn{1}{r|}{96.04} & \multicolumn{1}{r|}{36.30} & 37.18 & \multicolumn{1}{r|}{90.48} & \multicolumn{1}{r|}{\textbf{68.91}} & 72.30 \\ \hline
\multirow{3}{*}{\parbox[c]{0.7cm}{\centering $64$ $\times$ $64$}} & 10 & \multicolumn{1}{r|}{1.00} & \multicolumn{1}{r|}{0.99} & 0.99 & \multicolumn{1}{r|}{113.38} & \multicolumn{1}{r|}{144.54} & 137.14 & \multicolumn{1}{r|}{97.92} & \multicolumn{1}{r|}{15.00} & 16.43 & \multicolumn{1}{r|}{105.65} & \multicolumn{1}{r|}{79.77} & \textbf{76.79} \\
 & 20 & \multicolumn{1}{r|}{1.00} & \multicolumn{1}{r|}{0.97} & 0.93 & \multicolumn{1}{r|}{114.56} & \multicolumn{1}{r|}{156.03} & 141.26 & \multicolumn{1}{r|}{113.52} & \multicolumn{1}{r|}{33.24} & 28.34 & \multicolumn{1}{r|}{114.04} & \multicolumn{1}{r|}{94.64} & \textbf{84.80} \\
 & 30 & \multicolumn{1}{r|}{0.22} & \multicolumn{1}{r|}{0.98} & 0.90 & \multicolumn{1}{r|}{115.59} & \multicolumn{1}{r|}{142.65} & 145.51 & \multicolumn{1}{r|}{107.64} & \multicolumn{1}{r|}{57.31} & 61.43 & \multicolumn{1}{r|}{111.62} & \multicolumn{1}{r|}{\textbf{99.98}} & 103.47 \\ \hline
\end{tabular}%
}
\label{tab:num_agents}
\end{table}

\begin{table}[t]
\centering
\caption{Results for MFC-EQ and centralized baselines with different formation sizes in various sizes of grids.}
\Huge\scalebox{0.26}{%
\begin{tabular}{c|c|rrr|rrr|rrr|rrr}
\hline
 &  & \multicolumn{3}{c|}{Success Rate} & \multicolumn{3}{c|}{Makespan} & \multicolumn{3}{c|}{Form. Dev.} & \multicolumn{3}{c}{$\mathtt{MIX}(0.5)$} \\ \hline
\begin{tabular}[c]{@{}c@{}}Map\\ Size\end{tabular} & \begin{tabular}[c]{@{}c@{}}Form\\ Size\end{tabular} & \multicolumn{1}{c|}{\rotatebox[origin=c]{90}{\parbox[c]{2.3cm}{\centering SPP}}} & \multicolumn{1}{c|}{\rotatebox[origin=c]{90}{\parbox[c]{2.3cm}{\centering SWA- RM}}} & \multicolumn{1}{c|}{\rotatebox[origin=c]{90}{\parbox[c]{2.3cm}{\centering MFC- EQ}}} & \multicolumn{1}{c|}{\rotatebox[origin=c]{90}{\parbox[c]{2.3cm}{\centering SPP}}} & \multicolumn{1}{c|}{\rotatebox[origin=c]{90}{\parbox[c]{2.3cm}{\centering SWA- RM}}} & \multicolumn{1}{c|}{\rotatebox[origin=c]{90}{\parbox[c]{2.3cm}{\centering MFC- EQ}}} & \multicolumn{1}{c|}{\rotatebox[origin=c]{90}{\parbox[c]{2.3cm}{\centering SPP}}} & \multicolumn{1}{c|}{\rotatebox[origin=c]{90}{\parbox[c]{2.3cm}{\centering SWA- RM}}} & \multicolumn{1}{c|}{\rotatebox[origin=c]{90}{\parbox[c]{2.3cm}{\centering MFC- EQ}}} & \multicolumn{1}{c|}{\rotatebox[origin=c]{90}{\parbox[c]{2.3cm}{\centering SPP}}} & \multicolumn{1}{c|}{\rotatebox[origin=c]{90}{\parbox[c]{2.3cm}{\centering SWA- RM}}} & \multicolumn{1}{c}{\rotatebox[origin=c]{90}{\parbox[c]{2.3cm}{\centering MFC- EQ}}} \\ \hline
\multirow{3}{*}{\parbox[c]{0.7cm}{\centering $32$ $\times$ $32$}} & \LARGE{$7 \!\times 7$} & \multicolumn{1}{r|}{0.94} & \multicolumn{1}{r|}{1.00} & 0.97 & \multicolumn{1}{r|}{53.81} & \multicolumn{1}{r|}{66.40} & 68.30 & \multicolumn{1}{r|}{44.66} & \multicolumn{1}{r|}{12.37} & 9.06 & \multicolumn{1}{r|}{49.24} & \multicolumn{1}{r|}{39.39} & \textbf{38.68} \\
 & \LARGE{$9 \!\times 9$} & \multicolumn{1}{r|}{1.00} & \multicolumn{1}{r|}{1.00} & 1.00 & \multicolumn{1}{r|}{48.56} & \multicolumn{1}{r|}{63.20} & 67.33 & \multicolumn{1}{r|}{29.34} & \multicolumn{1}{r|}{9.10} & 8.72 & \multicolumn{1}{r|}{38.95} & \multicolumn{1}{r|}{\textbf{36.15}} & 38.03 \\
 & \LARGE{$11 \!\times 11$} & \multicolumn{1}{r|}{1.00} & \multicolumn{1}{r|}{1.00} & 1.00 & \multicolumn{1}{r|}{44.18} & \multicolumn{1}{r|}{57.75} & 55.12 & \multicolumn{1}{r|}{20.80} & \multicolumn{1}{r|}{7.03} & 8.32 & \multicolumn{1}{r|}{32.49} & \multicolumn{1}{r|}{32.39} & \textbf{31.72} \\ \hline
\multirow{3}{*}{\parbox[c]{0.7cm}{\centering $48$ $\times$ $48$}} & \LARGE{$7 \!\times 7$} & \multicolumn{1}{r|}{0.98} & \multicolumn{1}{r|}{1.00} & 0.87 & \multicolumn{1}{r|}{86.26} & \multicolumn{1}{r|}{110.94} & 107.37 & \multicolumn{1}{r|}{82.85} & \multicolumn{1}{r|}{19.84} & 22.40 & \multicolumn{1}{r|}{84.56} & \multicolumn{1}{r|}{65.39} & \textbf{64.89} \\
 & \LARGE{$9 \!\times 9$} & \multicolumn{1}{r|}{0.93} & \multicolumn{1}{r|}{0.96} & 0.93 & \multicolumn{1}{r|}{81.49} & \multicolumn{1}{r|}{109.67} & 98.64 & \multicolumn{1}{r|}{65.74} & \multicolumn{1}{r|}{21.03} & 16.84 & \multicolumn{1}{r|}{73.62} & \multicolumn{1}{r|}{65.35} & \textbf{57.74} \\
 & \LARGE{$11 \!\times 11$} & \multicolumn{1}{r|}{1.00} & \multicolumn{1}{r|}{1.00} & 1.00 & \multicolumn{1}{r|}{77.06} & \multicolumn{1}{r|}{105.06} & 97.26 & \multicolumn{1}{r|}{60.65} & \multicolumn{1}{r|}{15.12} & 14.08 & \multicolumn{1}{r|}{68.86} & \multicolumn{1}{r|}{60.09} & \textbf{55.67} \\ \hline
\multirow{3}{*}{\parbox[c]{0.7cm}{\centering $64$ $\times$ $64$}} & \LARGE{$7 \!\times 7$} & \multicolumn{1}{r|}{0.87} & \multicolumn{1}{r|}{0.99} & 0.96 & \multicolumn{1}{r|}{118.64} & \multicolumn{1}{r|}{155.36} & 138.84 & \multicolumn{1}{r|}{115.38} & \multicolumn{1}{r|}{31.07} & 55.42 & \multicolumn{1}{r|}{117.01} & \multicolumn{1}{r|}{\textbf{93.22}} & 97.13 \\
 & \LARGE{$9 \!\times 9$} & \multicolumn{1}{r|}{1.00} & \multicolumn{1}{r|}{1.00} & 0.96 & \multicolumn{1}{r|}{113.87} & \multicolumn{1}{r|}{153.33} & 133.92 & \multicolumn{1}{r|}{108.57} & \multicolumn{1}{r|}{25.95} & 33.20 & \multicolumn{1}{r|}{111.22} & \multicolumn{1}{r|}{89.64} & \textbf{83.56} \\
 & \LARGE{$11 \!\times 11$} & \multicolumn{1}{r|}{1.00} & \multicolumn{1}{r|}{0.95} & 1.00 & \multicolumn{1}{r|}{109.43} & \multicolumn{1}{r|}{149.61} & 131.08 & \multicolumn{1}{r|}{97.68} & \multicolumn{1}{r|}{23.02} & 27.75 & \multicolumn{1}{r|}{103.56} & \multicolumn{1}{r|}{86.32} & \textbf{79.42} \\ \hline
\end{tabular}%
}
\label{tab:form_size}
\end{table}

\begin{table}[t]
\centering
\caption{Results for MFC-EQ and centralized baselines for tests with a dynamic formation.}
\Huge\scalebox{0.26}{%
\begin{tabular}{c|rrr|rrr|rrr|rrr}
\hline
 & \multicolumn{3}{c|}{Success Rate} & \multicolumn{3}{c|}{Makespan} & \multicolumn{3}{c|}{Form. Dev.} & \multicolumn{3}{c}{$\mathtt{MIX}(0.5)$} \\ \hline
$M$ & \multicolumn{1}{c|}{\rotatebox[origin=c]{90}{\parbox[c]{2.3cm}{\centering SPP}}} & \multicolumn{1}{c|}{\rotatebox[origin=c]{90}{\parbox[c]{2.3cm}{\centering SWA- RM}}} & \multicolumn{1}{c|}{\rotatebox[origin=c]{90}{\parbox[c]{2.3cm}{\centering MFC- EQ}}} & \multicolumn{1}{c|}{\rotatebox[origin=c]{90}{\parbox[c]{2.3cm}{\centering SPP}}} & \multicolumn{1}{c|}{\rotatebox[origin=c]{90}{\parbox[c]{2.3cm}{\centering SWA- RM}}} & \multicolumn{1}{c|}{\rotatebox[origin=c]{90}{\parbox[c]{2.3cm}{\centering MFC- EQ}}} & \multicolumn{1}{c|}{\rotatebox[origin=c]{90}{\parbox[c]{2.3cm}{\centering SPP}}} & \multicolumn{1}{c|}{\rotatebox[origin=c]{90}{\parbox[c]{2.3cm}{\centering SWA- RM}}} & \multicolumn{1}{c|}{\rotatebox[origin=c]{90}{\parbox[c]{2.3cm}{\centering MFC- EQ}}} & \multicolumn{1}{c|}{\rotatebox[origin=c]{90}{\parbox[c]{2.3cm}{\centering SPP}}} & \multicolumn{1}{c|}{\rotatebox[origin=c]{90}{\parbox[c]{2.3cm}{\centering SWA- RM}}} & \multicolumn{1}{c}{\rotatebox[origin=c]{90}{\parbox[c]{2.3cm}{\centering MFC- EQ}}} \\ \hline
10 & \multicolumn{1}{r|}{1.00} & \multicolumn{1}{r|}{0.98} & 0.96 & \multicolumn{1}{r|}{48.19} & \multicolumn{1}{r|}{59.00} & 56.33 & \multicolumn{1}{r|}{127.90} & \multicolumn{1}{r|}{172.40} & 104.61 & \multicolumn{1}{r|}{88.05} & \multicolumn{1}{r|}{115.70} & \textbf{80.47} \\
15 & \multicolumn{1}{r|}{1.00} & \multicolumn{1}{r|}{1.00} & 1.00 & \multicolumn{1}{r|}{48.29} & \multicolumn{1}{r|}{63.85} & 57.56 & \multicolumn{1}{r|}{132.71} & \multicolumn{1}{r|}{210.82} & 114.33 & \multicolumn{1}{r|}{90.50} & \multicolumn{1}{r|}{137.34} & \textbf{85.95} \\
20 & \multicolumn{1}{r|}{0.97} & \multicolumn{1}{r|}{1.00} & 1.00 & \multicolumn{1}{r|}{49.64} & \multicolumn{1}{r|}{63.60} & 59.07 & \multicolumn{1}{r|}{141.18} & \multicolumn{1}{r|}{208.28} & 118.42 & \multicolumn{1}{r|}{95.41} & \multicolumn{1}{r|}{135.94} & \textbf{88.75} \\
25 & \multicolumn{1}{r|}{0.72} & \multicolumn{1}{r|}{1.00} & 1.00 & \multicolumn{1}{r|}{50.56} & \multicolumn{1}{r|}{62.58} & 61.29 & \multicolumn{1}{r|}{149.61} & \multicolumn{1}{r|}{204.33} & 123.20 & \multicolumn{1}{r|}{100.09} & \multicolumn{1}{r|}{133.46} & \textbf{92.25} \\
30 & \multicolumn{1}{r|}{0.90} & \multicolumn{1}{r|}{0.98} & 0.93 & \multicolumn{1}{r|}{50.00} & \multicolumn{1}{r|}{59.31} & 62.50 & \multicolumn{1}{r|}{146.82} & \multicolumn{1}{r|}{187.08} & 129.74 & \multicolumn{1}{r|}{98.41} & \multicolumn{1}{r|}{123.20} & \textbf{96.12} \\
35 & \multicolumn{1}{r|}{0.48} & \multicolumn{1}{r|}{0.94} & 0.87 & \multicolumn{1}{r|}{51.75} & \multicolumn{1}{r|}{57.87} & 64.71 & \multicolumn{1}{r|}{163.40} & \multicolumn{1}{r|}{189.64} & 133.07 & \multicolumn{1}{r|}{107.58} & \multicolumn{1}{r|}{123.76} & \textbf{98.89} \\
40 & \multicolumn{1}{r|}{0.25} & \multicolumn{1}{r|}{0.81} & 0.74 & \multicolumn{1}{r|}{52.68} & \multicolumn{1}{r|}{54.12} & 65.29 & \multicolumn{1}{r|}{168.64} & \multicolumn{1}{r|}{165.05} & 137.33 & \multicolumn{1}{r|}{110.66} & \multicolumn{1}{r|}{109.59} & \textbf{101.31} \\ \hline
\end{tabular}%
}
\label{tab:dynamic_formation}
\end{table}

\subsubsection{Dynamic Formation}
We tested these methods for more challenging tests where agents were required to adjust to different formations on the fly. Specifically, the desired formation changes to a different one at $T_{\emph{th}} = 30$. Centralized methods cannot handle such tasks effectively, as agents' paths must be planned before execution. In contrast, MFC-EQ allows each decentralized agent to be notified of the new formation, resulting in updated calculations of relative positions, which enables the agents to adjust to the new formation seamlessly. The results shown in Table~\ref{tab:dynamic_formation} suggest that MFC-EQ offers the flexibility needed to manage changing formations, whereas other methods result in significantly larger formation deviations.

\begin{figure}[t]
\centering
\includegraphics[width=\columnwidth]{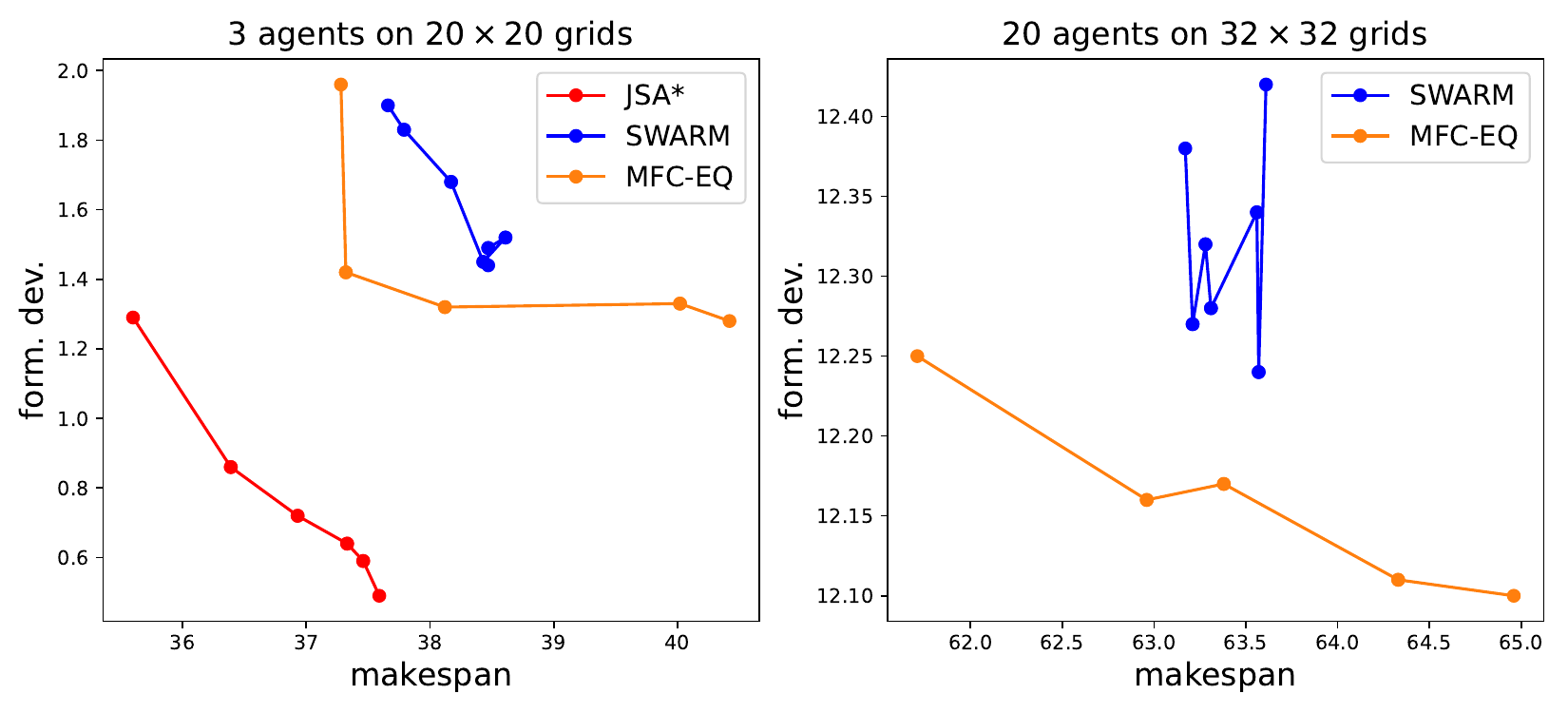}
\caption{Trade-off of makespan and formation deviation.}
\label{fig:trade_off}
\end{figure}

\subsubsection{Makespan and Formation Trade-off}
We further compare MFC-EQ with other methods across different preferences. Due to the limited scalability of JSA*, we first use instances with a $20 \times 20$ map size, $3 \times 3$ formation size, $15\%$ obstacle density, and $3$ agents. We vary $\epsilon$ from $1.0$ to $1.8$ for JSA*, $w$ from $1.0$ to $1.6$ for SWARM, and $\lambda$ (for $\bm{\omega}$) from $0.1$ to $0.9$ for MFC-EQ. JSA* can provide the Pareto-optimal frontier only for small-scale instances. We then repeat this experiment on larger instances with a $32 \times 32$ map size, $9 \times 9$ formation size, and $20$ agents. The results are shown in Fig.~\ref{fig:trade_off}. SPP produces solutions with near-optimal makespans but significantly larger formation deviations compared to the other methods. In large-scale cases, SWARM tends to fluctuate, meaning that even with more makespan allowance, it may result in worse formation deviations. MFC-EQ generates a near-convex envelope that encompasses all solutions from SWARM, although it remains suboptimal. Additionally, it offers a wider range of makespan options with greater solution variety.

\section{Conclusion and Future Work}
We proposed MFC-EQ, a general $Q$-learning framework for solving decentralized MAiF under conditions of partial observation and limited communication. MFC-EQ leverages mean-field approximation to simplify complex multi-agent interactions and employs envelop $Q$-learning to adapt to various preferences in this bi-objective task. Empirical results demonstrate that MFC-EQ outperforms existing centralized baselines in most scenarios and is particularly versatile and effective in handling dynamically changing desired formations. Moreover, MFC-EQ is not restricted to solving the decentralized MAiF and has significant potential for generalization to other multi-objective tasks in multi-agent systems.

While MFC-EQ shows strong performance in this initial attempt, there is certainly room for improvement, as indicated by the experimental results. One promising direction is to design a mixing network~\cite{rashid2020monotonic,hu2023mo} to estimate the joint action-value function for handling both objectives. Another direction is to apply other multi-objective RL algorithms that do not request the setting of linear preference, such as \cite{van2013scalarized,van2014multi,mannor2004geometric}. Regarding the agents' policy networks, integrating the current architecture with more advanced network designs~\cite{he2024alpha} and enhanced communication mechanisms~\cite{ma2021learning,li2022multi,wang2023scrimp} might further improve performance. Last but not least, we could also incorporate high-level global guidance as in~\cite{liu2020mapper,wang2020mobile} for both objectives. We leave these directions for future exploration.






\bibliographystyle{IEEEtran}
\bibliography{references}

\begin{thebibliography}{10}
\providecommand{\url}[1]{#1}
\csname url@rmstyle\endcsname
\providecommand{\newblock}{\relax}
\providecommand{\bibinfo}[2]{#2}
\providecommand\BIBentrySTDinterwordspacing{\spaceskip=0pt\relax}
\providecommand\BIBentryALTinterwordstretchfactor{4}
\providecommand\BIBentryALTinterwordspacing{\spaceskip=\fontdimen2\font plus
\BIBentryALTinterwordstretchfactor\fontdimen3\font minus
  \fontdimen4\font\relax}
\providecommand\BIBforeignlanguage[2]{{%
\expandafter\ifx\csname l@#1\endcsname\relax
\typeout{** WARNING: IEEEtran.bst: No hyphenation pattern has been}%
\typeout{** loaded for the language `#1'. Using the pattern for}%
\typeout{** the default language instead.}%
\else
\language=\csname l@#1\endcsname
\fi
#2}}

\bibitem{stern2019multi}
R.~Stern, N.~Sturtevant, A.~Felner, S.~Koenig, H.~Ma, T.~Walker, J.~Li,
  D.~Atzmon, L.~Cohen, T.~Kumar, \emph{et~al.}, ``Multi-agent pathfinding:
  Definitions, variants, and benchmarks,'' in \emph{SoCS}, 2019, pp. 151--158.

\bibitem{ma2017ai}
H.~Ma and S.~Koenig, ``Ai buzzwords explained: multi-agent path finding
  (mapf),'' \emph{AI Matters}, vol.~3, no.~3, pp. 15--19, 2017.

\bibitem{wurman2008coordinating}
P.~R. Wurman, R.~D'Andrea, and M.~Mountz, ``Coordinating hundreds of
  cooperative, autonomous vehicles in warehouses,'' \emph{AI magazine},
  vol.~29, no.~1, pp. 9--9, 2008.

\bibitem{morris2016planning}
R.~Morris, C.~S. Pasareanu, K.~S. Luckow, W.~Malik, H.~Ma, T.~S. Kumar, and
  S.~Koenig, ``Planning, scheduling and monitoring for airport surface
  operations.'' in \emph{AAAI Workshop: Planning for Hybrid Systems}, 2016, pp.
  608--614.

\bibitem{ma2017feasibility}
H.~Ma, J.~Yang, L.~Cohen, T.~Kumar, and S.~Koenig, ``Feasibility study: Moving
  non-homogeneous teams in congested video game environments,'' in
  \emph{AIIDE}, 2017, pp. 270--272.

\bibitem{gautam2012review}
A.~Gautam and S.~Mohan, ``A review of research in multi-robot systems,'' in
  \emph{ICIIS}, 2012, pp. 1--5.

\bibitem{li2020moving}
J.~Li, K.~Sun, H.~Ma, A.~Felner, T.~Kumar, and S.~Koenig, ``Moving agents in
  formation in congested environments,'' in \emph{SoCS}, 2020, pp. 131--132.

\bibitem{sartoretti2019primal}
G.~Sartoretti, J.~Kerr, Y.~Shi, G.~Wagner, T.~S. Kumar, S.~Koenig, and
  H.~Choset, ``Primal: Pathfinding via reinforcement and imitation multi-agent
  learning,'' \emph{IEEE Robotics and Automation Letters}, vol.~4, no.~3, pp.
  2378--2385, 2019.

\bibitem{liu2020mapper}
Z.~Liu, B.~Chen, H.~Zhou, G.~Koushik, M.~Hebert, and D.~Zhao, ``Mapper:
  Multi-agent path planning with evolutionary reinforcement learning in mixed
  dynamic environments,'' in \emph{IROS}, 2020, pp. 11\,748--11\,754.

\bibitem{ma2021distributed}
Z.~Ma, Y.~Luo, and H.~Ma, ``Distributed heuristic multi-agent path finding with
  communication,'' in \emph{ICRA}, 2021, pp. 8699--8705.

\bibitem{littman1994markov}
M.~L. Littman, ``Markov games as a framework for multi-agent reinforcement
  learning,'' in \emph{Machine learning proceedings}.\hskip 1em plus 0.5em
  minus 0.4em\relax Elsevier, 1994, pp. 157--163.

\bibitem{lin2023sacha}
Q.~Lin and H.~Ma, ``Sacha: Soft actor-critic with heuristic-based attention for
  partially observable multi-agent path finding,'' \emph{IEEE Robotics and
  Automation Letters}, vol.~8, no.~8, pp. 2377--3766, 2023.

\bibitem{liu2021moving}
S.~Liu, L.~Wen, J.~Cui, X.~Yang, J.~Cao, and Y.~Liu, ``Moving forward in
  formation: a decentralized hierarchical learning approach to multi-agent
  moving together,'' in \emph{IROS}, 2021, pp. 4777--4784.

\bibitem{stanley1971phase}
H.~E. Stanley, \emph{Phase transitions and critical phenomena}.\hskip 1em plus
  0.5em minus 0.4em\relax Clarendon Press, Oxford, 1971, vol.~7.

\bibitem{yang2018mean}
Y.~Yang, R.~Luo, M.~Li, M.~Zhou, W.~Zhang, and J.~Wang, ``Mean field
  multi-agent reinforcement learning,'' in \emph{ICML}, 2018, pp. 5571--5580.

\bibitem{Srirampomfrl2021}
S.~G. Subramanian, M.~E. Taylor, M.~Crowley, and P.~Poupart, ``Partially
  observable mean field reinforcement learning,'' in \emph{AAMAS}.\hskip 1em
  plus 0.5em minus 0.4em\relax IFAAMAS, 2021.

\bibitem{gabor1998multi}
Z.~G{\'a}bor, Z.~Kalm{\'a}r, and C.~Szepesv{\'a}ri, ``Multi-criteria
  reinforcement learning.'' in \emph{ICML}, 1998, pp. 197--205.

\bibitem{mannor2001steering}
S.~Mannor and N.~Shimkin, ``The steering approach for multi-criteria
  reinforcement learning,'' \emph{NeurIPS}, 2001.

\bibitem{natarajan2005dynamic}
S.~Natarajan and P.~Tadepalli, ``Dynamic preferences in multi-criteria
  reinforcement learning,'' in \emph{ICML}, 2005, pp. 601--608.

\bibitem{parisi2014policy}
S.~Parisi, M.~Pirotta, N.~Smacchia, L.~Bascetta, and M.~Restelli, ``Policy
  gradient approaches for multi-objective sequential decision making,'' in
  \emph{IJCNN}, 2014, pp. 2323--2330.

\bibitem{van2014multi}
K.~Van~Moffaert and A.~Now{\'e}, ``Multi-objective reinforcement learning using
  sets of pareto dominating policies,'' \emph{Journal of Machine Learning
  Research}, vol.~15, no.~1, pp. 3483--3512, 2014.

\bibitem{chen2019meta}
X.~Chen, A.~Ghadirzadeh, M.~Bj{\"o}rkman, and P.~Jensfelt, ``Meta-learning for
  multi-objective reinforcement learning,'' in \emph{IROS}, 2019, pp. 977--983.

\bibitem{castelletti2011multi}
A.~Castelletti, F.~Pianosi, and M.~Restelli, ``Multi-objective fitted
  q-iteration: Pareto frontier approximation in one single run,'' in
  \emph{ICNSC}, 2011, pp. 260--265.

\bibitem{abels2019dynamic}
A.~Abels, D.~Roijers, T.~Lenaerts, A.~Now{\'e}, and D.~Steckelmacher, ``Dynamic
  weights in multi-objective deep reinforcement learning,'' in \emph{ICML},
  2019, pp. 11--20.

\bibitem{yang2019generalized}
R.~Yang, X.~Sun, and K.~Narasimhan, ``A generalized algorithm for
  multi-objective reinforcement learning and policy adaptation,''
  \emph{NeurIPS}, 2019.

\bibitem{watkins1992q}
C.~J. Watkins and P.~Dayan, ``Q-learning,'' \emph{Machine learning}, vol.~8,
  pp. 279--292, 1992.

\bibitem{hasselt2010double}
H.~Hasselt, ``Double q-learning,'' \emph{NeurIPS}, 2010.

\bibitem{watson1989modern}
L.~T. Watson and R.~T. Haftka, ``Modern homotopy methods in optimization,''
  \emph{Computer Methods in Applied Mechanics and Engineering}, vol.~74, no.~3,
  pp. 289--305, 1989.

\bibitem{bengio2009curriculum}
Y.~Bengio, J.~Louradour, R.~Collobert, and J.~Weston, ``Curriculum learning,''
  in \emph{ICML}, 2009, pp. 41--48.

\bibitem{silver2005cooperative}
D.~Silver, ``Cooperative pathfinding,'' in \emph{AIIDE}, 2005, pp. 117--122.

\bibitem{sharon2015conflict}
G.~Sharon, R.~Stern, A.~Felner, and N.~R. Sturtevant, ``Conflict-based search
  for optimal multi-agent pathfinding,'' \emph{Artificial Intelligence}, vol.
  219, pp. 40--66, 2015.

\bibitem{pearl1982studies}
J.~Pearl and J.~H. Kim, ``Studies in semi-admissible heuristics,'' \emph{IEEE
  Transactions on Pattern Analysis and Machine Intelligence}, no.~4, pp.
  392--399, 1982.

\bibitem{haimes1971bicriterion}
Y.~Haimes, ``On a bicriterion formulation of the problems of integrated system
  identification and system optimization,'' \emph{IEEE Transactions on Systems,
  Man, and Cybernetics}, no.~3, pp. 296--297, 1971.

\bibitem{rashid2020monotonic}
T.~Rashid, M.~Samvelyan, C.~S. De~Witt, G.~Farquhar, J.~Foerster, and
  S.~Whiteson, ``Monotonic value function factorisation for deep multi-agent
  reinforcement learning,'' \emph{Journal of Machine Learning Research},
  vol.~21, no.~1, pp. 7234--7284, 2020.

\bibitem{hu2023mo}
T.~Hu, B.~Luo, C.~Yang, and T.~Huang, ``Mo-mix: Multi-objective multi-agent
  cooperative decision-making with deep reinforcement learning,'' \emph{IEEE
  Transactions on Pattern Analysis and Machine Intelligence}, 2023.

\bibitem{van2013scalarized}
K.~Van~Moffaert, M.~M. Drugan, and A.~Now{\'e}, ``Scalarized multi-objective
  reinforcement learning: Novel design techniques,'' in \emph{ADPRL}.\hskip 1em
  plus 0.5em minus 0.4em\relax IEEE, 2013, pp. 191--199.

\bibitem{mannor2004geometric}
S.~Mannor and N.~Shimkin, ``A geometric approach to multi-criterion
  reinforcement learning,'' \emph{JMLR}, vol.~5, pp. 325--360, 2004.

\bibitem{he2024alpha}
C.~He, T.~Yang, T.~Duhan, Y.~Wang, and G.~Sartoretti, ``Alpha: Attention-based
  long-horizon pathfinding in highly-structured areas,'' in \emph{ICRA}.\hskip
  1em plus 0.5em minus 0.4em\relax IEEE, 2024, pp. 14\,576--14\,582.

\bibitem{ma2021learning}
Z.~Ma, Y.~Luo, and J.~Pan, ``Learning selective communication for multi-agent
  path finding,'' \emph{IEEE Robotics and Automation Letters}, vol.~7, no.~2,
  pp. 1455--1462, 2021.

\bibitem{li2022multi}
W.~Li, H.~Chen, B.~Jin, W.~Tan, H.~Zha, and X.~Wang, ``Multi-agent path finding
  with prioritized communication learning,'' in \emph{ICRA}.\hskip 1em plus
  0.5em minus 0.4em\relax IEEE, 2022, pp. 10\,695--10\,701.

\bibitem{wang2023scrimp}
Y.~Wang, B.~Xiang, S.~Huang, and G.~Sartoretti, ``Scrimp: Scalable
  communication for reinforcement-and imitation-learning-based multi-agent
  pathfinding,'' in \emph{IROS}, 2023, pp. 9301--9308.

\bibitem{wang2020mobile}
B.~Wang, Z.~Liu, Q.~Li, and A.~Prorok, ``Mobile robot path planning in dynamic
  environments through globally guided reinforcement learning,'' \emph{IEEE
  Robotics and Automation Letters}, vol.~5, no.~4, pp. 6932--6939, 2020.

\end{thebibliography}

\end{document}